\documentclass[letterpaper, 10 pt, conference]{ieeeconf}  

\IEEEoverridecommandlockouts                              

\usepackage{algorithmicx}
\usepackage{algorithm}
\usepackage{algpseudocode}
\usepackage{amsfonts}
\usepackage{amsmath}
\usepackage{amssymb}
\usepackage{multicol}
\usepackage[bookmarks=true]{hyperref}
\usepackage{bm}
\usepackage{booktabs}
\usepackage{graphics} 
\usepackage{graphicx}
\usepackage{placeins}
\usepackage{svg}
\usepackage{subcaption}
\usepackage{tikz}
\usepackage{xcolor}
\usepackage{xspace}
\usepackage{xr}
\usepackage{times}

\makeatletter
\DeclareRobustCommand\onedot{\futurelet\@let@token\@onedot}
\def\@onedot{\ifx\@let@token.\else.\null\fi\xspace}

\def\aka{\emph{a.k.a}\onedot}
\def\eg{\emph{e.g}\onedot} \def\Eg{\emph{E.g}\onedot}

\def\etal{\emph{et al}\onedot}

\newcommand*{\transpose}{%
    {\mathpalette\@transpose{}}%
}
\newcommand*{\@transpose}[2]{%
    \raisebox{\depth}{$\m@th#1\intercal$}%
}
\makeatother

\newcommand{\HybridFactor}{\phi^\textrm{H}}
\newcommand{\HybridComponent}{\phi^m_\textrm{hf}}
\newcommand{\HybridNonlinearFactor}{\phi^\textrm{HNF}}

\newcommand{\HybridGFComponent}{\phi^m_\textrm{hgf}}

\newcommand{\HybridLinFactor}{\phi^\textrm{HG}}

\newcommand{\CC}{\mathcal{C}}
\newcommand{\MM}{\mathcal{M}}

\newcommand{\cS}{\mathcal{S}}  

\newcommand{\DMRThreshold}{\delta_{\text{DMR}}}

\newcommand{\Dim}{\textrm{dim}}

\newcommand{\SEThree}{\mathrm{SE}(3)}

\title{\LARGE \bf
Variable Elimination in Hybrid Factor Graphs for Discrete-Continuous Inference \& Estimation
}

\author{Varun Agrawal$^{1}$, Frank Dellaert$^{1}$
    \thanks{$^{1}$Institute for Robotics and Intelligent Machines and School of Interactive Computing, Georgia Institute of Technology, Atlanta, GA 30332 USA.
        {\tt\small \{varunagrawal,frank.dellaert\}@gatech.edu}}%
}

\begin{document}

\maketitle
\thispagestyle{empty}
\pagestyle{empty}

\begin{abstract}
Many problems in robotics involve both continuous and discrete components, and modeling them together for estimation tasks has been a long standing and difficult problem.
Hybrid Factor Graphs give us a mathematical framework to model these types of problems, however existing approaches for solving them are based on approximations.
In this work, we propose a new framework for hybrid factor graphs along with a novel variable elimination algorithm to produce a hybrid Bayes network, which can be used for exact \textit{Maximum A Posteriori} estimation and marginalization over both sets of variables.
Our approach first develops a novel hybrid Gaussian factor which can connect to both discrete and continuous variables, and a hybrid conditional which can represent multiple continuous hypotheses conditioned on the discrete variables.
Using these representations, we derive the process of hybrid variable elimination under the Conditional Linear Gaussian scheme, giving us exact posteriors as a hybrid Bayes network.
To bound the number of discrete hypotheses, we use a tree-structured representation of the factors coupled with a simple pruning and probabilistic assignment scheme, which allows for tractable inference.
We demonstrate the applicability of our framework on a large scale SLAM dataset and a real world pose graph optimization problem, both with ambiguous measurements which require discrete choices to be made for the most likely measurements. Our demonstrated results showcase the accuracy, generality, and simplicity of our hybrid factor graph framework.
\end{abstract}

\section{Introduction}\label{sec:intro}

Robotics inherently has both discrete and continuous components, making it vital to be able to model both.
Tasks such as Simultaneous Localization and Mapping (SLAM), grasping, Task and Motion Planning (TAMP), fault assessment~\cite{Lerner00aaai}, and locomotion (amongst others) all involve some discrete components along with an optimization task over continuous variables.
\Eg in SLAM, the continuous components are the positions of the robot and the landmarks, but the data association between features is a discrete estimation problem~\cite{Hsiao19icra,Doherty22ral}.
For manipulation, estimating the pose of the object to grasp, as well as identifying the discrete set of affordances and contacts on the object constitutes a hybrid estimation problem~\cite{Mahler17rss,Florence18corl,Suresh24scirobotics}, while in legged locomotion, being able to estimate discrete foot contact can aid with continuous state estimation via the kinematics~\cite{Camurri20frontiers_pronto}.
Moreover, since probabilistic methods are currently the \textit{de facto} approach for estimation\footnote[2]{The Kalman filter can also be viewed from a Bayesian perspective~\cite{Pei19arxiv}.}, with factor graphs and Bayes networks~\cite{Dellaert21ar} seeing significant recent success, we would prefer to apply similar insights to these \textit{hybrid} problems as well.

While factor graphs and variable elimination are well understood for purely continuous systems (\eg Gaussian SLAM) and purely discrete graphical models~\cite{Kschischang01it,Dellaert17fnt_fg}, robotics problems often require inference in hybrid models containing both discrete and continuous variables.
Existing factor graph frameworks either approximate these interactions or require specialized solvers~\cite{Oh05aaai,Roy07hybridsystems,Dong20icml,Doherty22ral}, while
more recent exact methods~\cite{Hsiao19icra,Jiang21arxiv} are limited in their modeling capabilities to specific problems and scenarios.


\begin{figure}[t]
    \centering
    \includegraphics{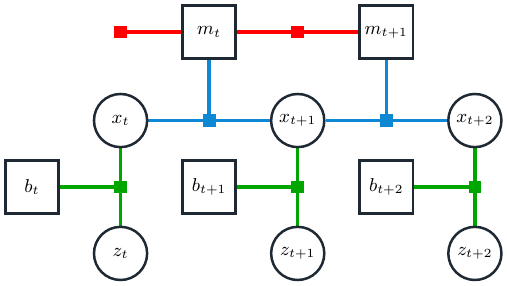}
    \caption{A hybrid factor graph for a state estimation problem with ambiguous motion models (blue factors), measurement models (green factors), and a discrete markov chain (red factors). $x$ is robot state, $z$ is measurement, and square nodes $m$ and $b$ are discrete switching modes.}
    \label{fig:hybrid_factor_graph}
\end{figure}

In this work we develop a principled extension of factor graph variable elimination to hybrid models under the Conditional Linear Gaussian (CLG) assumption.
To this end, we also provide novel definitions of \textit{hybrid factors} which can be used to specify all manner of relationships between discrete and continuous variables.
This allows us to model a wide variety of mixed discrete-continuous estimation problems, while computing exact posterior distributions as a hybrid Bayes network, subsuming existing approaches under a general approach.
\Eg a switching linear dynamical system can be easily represented using our framework, as depicted in Fig.~\ref{fig:hybrid_factor_graph}.
Similarly, we can model Mixture-of-Gaussians, multi-hypothesis SLAM (shown in section \ref{sec:experiments}), and others.

We first define our hybrid factor graphs, with both linear and nonlinear variants, going beyond existing definitions in the literature. 
Furthermore, our design considerations make them computationally efficient and easy to implement.
We then derive hybrid versions of both the Sum-Product and Max-Product algorithms as instances of the general variable elimination algorithm for \textit{Maximum A Posteriori} (MAP) estimation and marginal computation respectively.
Furthermore, we derive the set of components needed to bridge the gap between continuous and discrete variables, allowing us to extend existing ideas and tools~\cite{Murphy02phd} to the hybrid case, while still being natural to postulate and efficient to solve.
Additionally, we propose two probabilistic techniques for handling potentially exponential hypothesis growth and managing data storage for large scale problems.
By analyzing the complexity of hybrid inference, we show how hypothesis pruning and probabilistic modeling prove vital to efficient performance.
Finally, we experimentally validate our framework on two large scale, nonlinear tasks: SLAM and pose graph optimization problem, both of which consist of multiple ambiguous measurements and loop closures, and demonstrate improved performance over baselines.

Overall, our contributions are the following
\begin{itemize}
    \item A factor graph framework with novel hybrid factors for modeling discrete and continuous variables jointly.
    \item Hybrid Sum-Product and Max-Product algorithms for efficient variable elimination and MAP optimization.
    \item Complexity analysis of our approach along with probabilistic techniques to constrain the exponential growth of hypotheses during inference.
    \item Validation on a large-scale, nonlinear SLAM dataset and a real-world pose graph optimization problem, both with ambiguous odometry measurements and loop closures.
\end{itemize}

\section{Related Work}
\label{sec:related-work}

\subsection{Discrete-Continuous Optimization}

Discrete-Continuous optimization has been an active area of research for many years.
Most attempts at this kind of hybrid optimization fall into the realm of Mixed-Integer Programming (MIP) and Mixed-Integer Nonlinear Programming (MINLP)~\cite{Kannan94jmd_LagrangeMultiplierMIP}.
\cite{Grossman13aiche_gdp} detailed a new framework titled Generalized Disjunctive Programming (GDP), which is based on MIP, allowing for derivations of mixed-integer optimization problems with stronger continuous relaxations~\cite{Vecchietti03cce_gdp}.
Evolutionary strategies have also been explored in hybrid spaces, such as Genetic Algorithms~\cite{Ndiritu99eo} and differential evolutionary algorithms~\cite{Lampinen00report} for handling hybrid optimization.

Additionally, hybrid estimation has been widely studied in the field of state estimation for discrete-continuous or switching systems.
Its initial development was fueled by the need for tracking a variety of aircraft trajectories via Interacting Multiple Models and its variants~\cite{Blom88tac,Cox96tpami,Hwang06cta}. 
Recent work has also focused on estimation in Switching Dynamical Systems, both linear~\cite{Oh05aaai} and nonlinear~\cite{Dong20icml}.

Hybrid estimation has also seen applications in other fields. In computer vision, it has been used for depth and plan normal estimation from images~\cite{Liu14cvpr}, dense stereo matching~\cite{Shekhovtsov16arxiv}, Structure-from-Motion as discrete-continuous optimization~\cite{Crandall12pami}, optical flow~\cite{Roth09sgavma}, and visual multi-target tracking~\cite{Andriyenko12cvpr}. These applications use approximate methods or optimize each component independently rather than jointly.
Additionally, hybrid inference has also been explored in probabilistic programming languages~\cite{Wu18icml_hybrid_ppl}.

\subsection{Hybrid Inference}


Probabilistic inference over hybrid graphical models has received significant interest. Lauritzen \etal~\cite{Lauritzen95jasa} first developed exact inference over hybrid graphical models, defining the Conditional Linear Gaussian (CLG) as a class of models where discrete variables could not have continuous parents.
\cite{Lerner01uai} extended it by allowing discrete variables to have continuous parents, an approach we hope to adopt in future work.
\cite{Mori16esa} proposed to use decision-tree structured conditional probability tables for efficient inference over large discrete and continuous domains.
The influential work of~\cite{Murphy02phd} introduced Switching Kalman Filters~\cite{Murphy98report_switchingKF} and Dynamic Bayes Networks~\cite{Murphy01nips} as unified approaches to modeling hybrid problems in a Bayesian framework, and algorithms for inferring desired posteriors.

Various approaches to factor graph based hybrid inference have been proposed.
\cite{Stender21bhi_hybrid_fg} first defined the notion of a hybrid factor graph for biomedical data analysis but in a highly restrictive setting, using an iterative message passing approach to optimize for the MAP.
In contrast, our contribution is the development of both hybrid Bayes networks and hybrid factor graphs to follow the paradigm laid out by~\cite{Dellaert17fnt_fg}, which goes from a generative Bayes network to a factor graph after conditioning on measurements, and then uses the elimination algorithm to transform the factor graph back to a Bayes network representing the now hybrid posterior.

A multi-hypothesis extension to iSAM2~\cite{Kaess12ijrr} was proposed by~\cite{Hsiao19icra}, which only allows 3 types of hybrid relationships and requires the use of additional data structures for tracking the multiple hypotheses. Additionally, they do not support discrete probability relationships in their framework.
Similarly, \cite{Jiang21arxiv} use a tree-based structure to model incremental smoothers, with each branch corresponding to a single discrete assignment.
Our work both generalizes and simplifies these approaches, allowing for arbitrary types of hybrid relationships without the need for dedicated data structures to track multiple hypotheses.

\cite{Davey07tro} proposed using the Probabilistic Multi-Hypothesis Tracker (PMHT) for the SLAM problem with unknown data association, while \cite{Doherty22ral} developed an alternating optimization scheme by fixing either the discrete or continuous variables while updating the other.
Both of these methods require good initial continuous estimates and risk falling into an incorrect local minima due to inaccurate initial estimates. We avoid this by optimizing both parts jointly.


\section{Hybrid Bayes Networks \& Factor Graphs}

\subsection{Hybrid Bayes Network}\label{sec:hybrid-bayes-network}

In this work, we are interested in hybrid problems which can be modeled using a \textit{Hybrid Bayes Network} \cite{Lerner01uai}.
Concretely, we want to represent a hybrid joint probability distribution $P(X, M, Z)$, consisting of continuous states $X$, discrete modes $M$, and measurements $Z$, using a hybrid Bayes network, allowing us to perform probabilistic inference on systems of both continuous and discrete variables.

A hybrid Bayes network is a directed, acyclic graph where each node is a conditional probability distribution of a variable $a_j$ on its parents $\Pi_j$.
\begin{equation}
    P(X, M, Z) = \prod_{a_i \in X \cup M \cup Z} P(a_i | \Pi_i)
\end{equation}
This allows us to perform two tasks which are of particular interest to roboticists:
\begin{enumerate}
    \item Compute the posterior distribution $P(X, M | Z)$ of the variables $X, M$ given the measurements $Z$:
        \begin{equation}\label{eqn:posterior_probability}
            P(X, M \vert Z) \propto P(X, M, Z)
        \end{equation}
    \item Infer the \textit{Maximum A Posteriori} estimate of $X$ and $M$
    \begin{equation}\label{eqn:hybrid_map}
        X^*, M^* = \arg\max_{X, M}P(X, M \vert Z)
    \end{equation}
\end{enumerate}

\begin{figure}[h]
    \centering
    \includegraphics{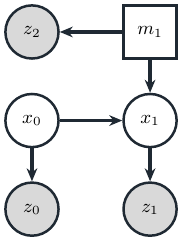}
    \caption{A hybrid Bayes network of a simple switching system, with 2 continuous latent variables $x_0, x_1$, 1 latent discrete variable $m_1$, and three measurements $z_0, z_1, q_0$.}
    \label{fig:simple_hybrid_bn}
\end{figure}

Hybrid Bayes networks can model every combination of parent-child relationships of the variables in the conditionals, between both discrete and continuous.
Fig.~\ref{fig:simple_hybrid_bn} illustrates a simple switching dynamical system~\cite{Murphy98report_switchingKF} using a HBN.


Every node $P(a_i | \Pi_i)$ in a hybrid Bayes network is a \textit{Hybrid Conditional}~\cite{Murphy02phd}, which is a set of conditional components $p^m(a_i | X_i)$ over continuous variables $X_i$ indexed by a specific assignment $m\in D(M_i) \subseteq M$ of the discrete variables.
Here $D(M) = (D_1 \times \ldots \times D_{|M|})$ is the domain function of the discrete variables within set $M$~\cite{Darwiche09book}.

For example, a Gaussian noise model can be specifed as a set of continuous components using a \textit{Gaussian density function}, indexed by the discrete variable.
Thus a hybrid conditional for a sensor model with multiple, possibly nonlinear functions $h^m(z; X)$ is defined as
\begin{align}
    \begin{split}
        &P(z | X_i, M_i) = \{ p^m(z | X_i); m \in D(M_i) \} \\
        &p^m(z | X_i) = \frac{1}{\sqrt{|2 \pi \Sigma^m|}} \exp \Big( -\frac{1}{2} \Vert z - h^m(X_i) \Vert^2_{\Sigma^m} \Big)
    \end{split}
\end{align}
Analogously, we can similarly define other types of conditionals, \eg for robust noise models.

\subsection{Hybrid Factor Graph}\label{sec:hybrid-factor-graph}

Including measurements as variables in hybrid Bayes networks significantly increases the state space size and makes it computationally more expensive to perform inference.
Since we have known measurements, we can model only the unknown state variables by following the approach in \cite{Dellaert17fnt_fg} and converting the Bayes network to a factor graph by conditioning on the measurements, but in a hybrid sense.

\begin{figure}[ht]
    \centering
    \includegraphics{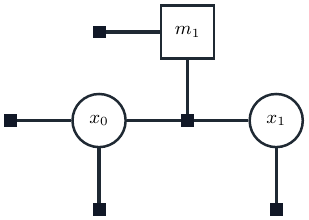}
    \caption{A simple hybrid factor graph with 2 continuous variables $x_0, x_1$ and 1 discrete variable $m_1$.}
    \label{fig:simple_hybrid_fg}
\end{figure}

Following existing works~\cite{Doherty22ral,Stender21bhi_hybrid_fg}, we use a \textit{Hybrid Factor Graph} where only the \textit{unknown} variables are retained.
A hybrid factor graph is defined as a product of hybrid factors over continuous variables $X$ and discrete variables $M$
\begin{equation}
    \Phi(X, M) = \prod_{\phi \in \Phi} \phi(\mathcal{X}, \mathcal{M})
\end{equation}
where each factor $\phi$ is defined over a subset of the variables $\mathcal{X} \subseteq X$, $\mathcal{M} \subseteq M$.
Fig.~\ref{fig:simple_hybrid_fg} showcases a hybrid factor graph converted from the hybrid Bayes network in Fig.~\ref{fig:simple_hybrid_bn}.


\section{Hybrid Variable Elimination}\label{sec:hybrid-elimination}

A major contribution of this work is a \textit{hybrid variable elimination algorithm} to convert a hybrid factor graph to a hybrid Bayes network representing the desired posterior distribution.
Similar to the continuous-only case~\cite{Kschischang01it,Dellaert17fnt_fg}, while a factor graph allows us to model our joint distribution conditioned on observed measurements, a Bayes network lets us perform various operations \eg MAP estimation, sampling, and marginalization.
Thus an efficient variable elimination algorithm for a hybrid factor graph is required for strong performance in robotics applications.

To that end, we first discuss the structure and assumptions of our hybrid factors, which differs from existing definitions and provides the necessary groundwork for our proposed algorithm.
These considerations form another key contribution.

\subsection{Hybrid Factor}\label{sec:hybrid-factor}

We define a \textit{Hybrid Factor} $\HybridFactor$ as a set of continuous factor components $\HybridComponent$, each indexed by discrete mode $m$.
Since we convert a hybrid Bayes network to a hybrid factor graph by conditioning on the known measurements, each hybrid factor component $\HybridComponent$ is defined using a likelihood function $\mathcal{L}$ obtained from the corresponding hybrid conditional.
\begin{equation}
    \HybridFactor(X_i, M_i) = \{ \HybridComponent(X_i) = \mathcal{L}^m(X_i; z); m \in D(M_i) \}
\end{equation}
Without loss of generality, we assume each component function $h^m: \mathbb{R}^{\Dim(x)} \to \mathbb{R}^{\Dim(\mu^m)}$ is linear with zero-mean Gaussian noise $\epsilon \in \mathcal{N}(0, \Sigma)$ over $x$, a vectorization of $X$
\begin{equation}
    h^m(x) = A^m x + \epsilon
\end{equation}
Each hybrid factor component $\HybridComponent$ is then modeled using a Gaussian density over the function, with a mode-dependent mean $\mu^m$ and covariance $\Sigma^m$.
This gives us a \textit{Hybrid Gaussian Factor} $\HybridLinFactor$, with $\HybridGFComponent$ components.
\begin{align}
    \begin{split}
        & \HybridLinFactor(X_i, M_i) = \Big\{ \HybridGFComponent(X_i); m \in D(M_i) \Big\} \\
        & \HybridGFComponent(X_i) = \frac{1}{\sqrt{|2\pi \Sigma^m|}} \exp \Big( -\frac{1}{2} \Vert A^m x_i - \mu^m \Vert^2_{\Sigma^m} \Big)
    \end{split}
\end{align}
As a concrete example, going forward we will use the mode-dependent observation model $z^m = H^m x + \epsilon$, with similar definitions for other types of models.

\subsection{Including the Normalizer}
Since the Gaussian PDF normalizer is potentially dependent on the discrete mode, it is imperative to account for it in the hybrid factor definition.
This consideration is a significant departure from existing works where the normalizer is ignored, and is a requirement for ensuring correctness of our proposed variable elimination algorithm.
\begin{align}
    \begin{split}
        &\HybridGFComponent(X_i) = \frac{1}{\sqrt{|2\pi \Sigma^m|}} \exp \Big( -\frac{1}{2} \Vert H^m x_i - z^m \Vert^2_{\Sigma^m} \Big) \\
        &= \exp \bigg( -\frac{1}{2} \Vert H^m x_i - z^m \Vert^2_{\Sigma^m} - \log(\sqrt{|2\pi \Sigma^m|}) \bigg)
    \end{split}
\end{align}
Taking the negative log-likelihood of $\HybridGFComponent$ gives us the square root form~\cite{Dellaert05rss}
\begin{align}\label{eqn:hybrid-factor-2}
    \begin{split}
        -\log &\HybridGFComponent(X_i) = \frac{1}{2} \Vert H x_i - z^m \Vert^2_{\Sigma^m} + \log(\sqrt{|2\pi \Sigma^m|}) \\
        &= \frac{1}{2} \bigg( \Vert A^m x_i - b^m \Vert^2_2 + 2 \log(\sqrt{|2\pi \Sigma^m|}) \bigg)
    \end{split}
\end{align}
with $A^m = (\Sigma^m)^{-1/2}H$, $b^m = (\Sigma^m)^{-1/2}z^m$.
We further simplify the factor component definition by using the following form which can be directly used in variable elimination.
\begin{align}
    \begin{split}
        &-\log \HybridGFComponent(X_i) = \frac{1}{2} \bigg( \Vert \tilde{A}^m x_i - \tilde{b}^m \Vert^2_2 \bigg) \\
        & \tilde{A}^m = \begin{bmatrix}A^m \\ 0 \end{bmatrix}, \tilde{b}^m = \begin{bmatrix}b^m \\ \sqrt{B} \end{bmatrix} \\
        &B = 2\log(\sqrt{|2 \pi \Sigma^m|}) - \min_{\tilde{m}} {2\log(\sqrt{|2\pi\Sigma^{\tilde{m}}} | )}
    \end{split}
\end{align}
where $B$ is defined in a way which ensures it is always positive. Please see the supplementary for more details.

\subsection{Representational Efficiency}

\begin{figure}[h]
    \centering
    \includegraphics{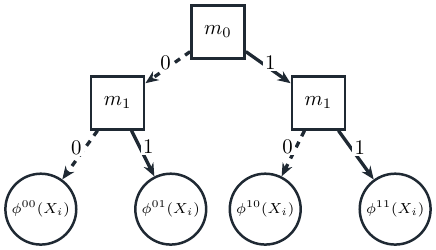}
    \caption{An example of the tree-structure for a hybrid factor $\phi(X_i,M_i)$ where $M_i = \{m_0, m_1\}$ and cardinality $\vert m_i \vert = 2$.}
    \label{fig:decision_tree}
\end{figure}

For computational efficiency, we store the components of a factor in a tree structure, as depicted in Fig.~\ref{fig:decision_tree}.
Each node in a level corresponds to a discrete variable with the cardinality providing the branching factor.
A tree data structure not only allows us to access each component in $O(\log M_i)$ time, but also allows for easy hypothesis pruning (Section~\ref{sec:pruning}).


\subsection{Variable Ordering}

\begin{figure}[h]
    \centering
    \includegraphics{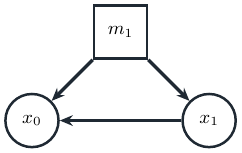}
    \caption{The hybrid Bayes network resulting from our variable elimination algorithm applied to the factor graph in Fig.~\ref{fig:simple_hybrid_fg}.}
    \label{fig:example_hybrid_bn}
\end{figure}

The variable elimination algorithm requires an ordering of the variables as input.
For hybrid factor graphs, we leverage the \textit{Conditional Linear Gaussian} (CLG) scheme~\cite{Lauritzen95jasa} which simplifies the algorithm derivation and allows us to perform exact inference while being widely applicable to a variety of problems.
The CLG scheme specifies a strong ordering~\cite[B.3.6]{Murphy02phd}, where all continuous variables are eliminated before the discrete variables, and results in a hybrid Bayes network where discrete variables cannot have continuous parents, \eg Fig.~\ref{fig:example_hybrid_bn}.
This leads to an efficient tree structure which, as we will see later, allows for better hypothesis management on large scale problems.
The ordering in-between only continuous or discrete variables is application specific.

In the following discussion, we derive hybrid versions of the Sum-Product and Max-Product algorithms as instances of the variable elimination algorithm for linear Gaussian factor graphs.
Particularly, we show how to handle the continuous-discrete boundary which arises due to our strong ordering, allowing for a consistent algorithm in both cases.
We limit our discussion to linear Gaussian factor graphs, leaving extensions to other types as future avenues of research.

\subsection{Sum-Product Algorithm}

The Sum-Product algorithm primarily involves a summation operation over the product of factors related to the variable being eliminated, resulting in hybrid conditionals which form the hybrid Bayes network representing the desired posterior over the latent variables $X, M$.
Given our variable ordering $\mathcal{O}$, we eliminate each variable $j \in \mathcal{O}$ in turn to obtain the conditional $p(j | \cS_j)$ and the separator factor $\tau(\cS_j)$ where $\cS_j = (\CC_j, \MM_j)$. Here, $\CC_j$ and $\MM_j$ correspond to the continuous and discrete separators, respectively.

Starting with a continuous variable $x_j \in X$ in our strong ordering, using all factors connected to $x_j$ we form a product factor $\psi(x_j, \cS_j)$ on corresponding discrete assignments.
\begin{align}
    \begin{split}
        \psi(x_j, \cS_j) &= \big\{ \psi^m(x_j, \CC_j); m \in D(\MM_j) \big\} \\
        \psi^m(x_j, \CC_j) &= \prod_{\phi_i \in \Phi} \phi^m_i(x_j, \CC_j) \\
        &= \exp \Big( - \sum_j \frac{1}{2}\Vert A^m_j \begin{bmatrix} x_j \\ \CC_j \end{bmatrix} -b^m_j \Vert^2_2 \Big)
    \end{split}
\end{align}
To support eliminating $x_j$, we factorize each product factor component $\psi^m$ into two parts, one involving $x_j$ and the other involving only the continuous separator $\CC_j$.
\begin{align}
    \begin{split}
        \psi^m(x_j, \CC_j) = &\exp \Big(-\frac{1}{2} \Vert R^m_j x_j + T^m_j \CC_j - d^m_j \Vert^2_2 \Big) \\
        & \exp \Big(-\frac{1}{2} \Vert A^m_{\tau}\CC_j - b^m_{\tau} \Vert^2_2 \Big)
    \end{split}
\end{align}
We sum over each product factor component $\psi^m$, resulting in the separator factor components $\tau^m(\CC_j)$, which is then used to obtain $m$-indexed conditional distributions $p^m(x_j | \CC_j)$.
\begin{align}\label{eqn:hybrid-sum-product}
    \begin{split}
        &\tau^m(\CC_j) = \int_{x_j} \psi^m(x_j, \CC_j) dx_j \\
        &= \exp \Big( -\frac{1}{2} \Vert A^m_{\tau}\CC_j - b^m_{\tau} \Vert^2_2 \Big) \Big(\frac{1}{\sqrt{|2\pi \Sigma^m|}}\Big)^{-1} \\
        &p^m(x_j | \CC_j) = \psi^m(x_j, \CC_j) / \tau^m(\CC_j) \\
        &= \frac{1}{\sqrt{|2\pi \Sigma^m|}}\exp \Big(-\frac{1}{2} \Vert R^m_j x_j + T^m_j \CC_j - d^m_j \Vert^2_2 \Big)
    \end{split}
\end{align}
When we eliminate the final continuous variable $x_n$, we will be at the \textit{continuous-discrete boundary}, which requires special care.
The product factor in this case is
\begin{align}
    \begin{split}
        &\psi(x_n, \MM_n) = \big\{ \psi^m(x_n); m \in D(\MM_n) \big\} \\
        &\psi^m(x_n) = \exp \Big( -\frac{1}{2} \Vert A^m x_n - b^m\Vert^2_2 \Big) \\
        &= \exp \bigg( -\frac{1}{2} \Vert R^m_n x_n - d^m_n \Vert^2_2 \bigg) \exp \bigg(-\frac{1}{2} \Vert b^m_{\tau} \Vert^2_2 \bigg)
    \end{split}
\end{align}
Here, the conditional $p(x_j | \MM_j)$ obtained is similar as before.
The separator factor $\tau(\MM_j)$ represents the leftover continuous residuals for each discrete index, and is a discrete factor.
\begin{equation}
    \tau^m() = \exp \bigg( -\frac{1}{2} \Vert b^m \Vert^2_2 - \log \Big( \frac{1}{\sqrt{|2\pi \Sigma^m|}} \Big) \bigg)
\end{equation}
At this point, we are left with only discrete factors due to our strong ordering. We can now compute the conditional distributions $P(m_j | \MM_j)$ and separator factors $\tau(\MM_j)$ as
\begin{align}\label{eqn:discrete-sum-product}
    \begin{split}
        & \tau(\MM_j) = \sum_{m_j} \psi(m_j, \MM_j) \\
        & P(m_j | \MM_j) = \psi(m_j, \MM_j) / \tau(\MM_j)
    \end{split}
\end{align}
This sequence of operations applied to each variable results in a set of conditionals, the product of which represents our desired hybrid Bayes network $P(X, M)$.

\subsection{Max-Product Algorithm}

While the Sum-Product algorithm gives us a Bayes network representing the posterior distribution, the Max-Product algorithm lets us estimate the \textit{MAP} of the unknown variables $X, M$.
Applying Max-Product over the hybrid factor graph results in a series of functions which we can query to get the MAP value for each variable.

Similar to Sum-Product, for each variable $x_j$ in our strong ordering, we form the product factor $\psi(x_j, \cS_j)$ of the factors connected to it.
Each product factor component $\psi^m(x_j, \CC_j)$ is then factorized into a function $g_j^m(\CC_j)$ of the maximum value of $x_j$ given the separator values and the separator factor $\tau^m(\CC_j)$, both indexed by the discrete variables.
\begin{align}\label{eqn:hybrid-max-product}
    \begin{split}
        & g_j(\CC_j, \MM_j) = \big\{ g^m_j(\CC_j); m \in D(\MM_j) \big\} \\
        & g^m_j(\CC_j) = \arg\max_{x_j} \psi^m(x_j, \CC_j) = (R^m_j)^{-1}(d^m_j - T^m_j \CC_j) \\
        & \tau(\CC_j, \MM_j) = \big\{ \tau^m(\CC_j); m \in D(\MM_j) \big\} \\
        & \tau^m(\CC_j) = \max_{x_j} \psi^m(x_j, \CC_j) = \exp \Big(-\frac{1}{2}\Vert A^m_{\tau}\CC_j - b^m_{\tau} \Vert^2_2 \Big)
    \end{split}
\end{align}
On reaching the \textit{continuous-discrete boundary} on eliminating the final continuous variable $x_n$, we similarly obtain the separator factor as a discrete factor on the residuals:
\begin{align}\label{eqn:discrete-max-product}
    \begin{split}
    & g^m_n() = \arg\max_{x_n} \psi^m(x_n, \MM_n) = (R^m_n)^{-1} d^m_n \\
    & \tau^m() = \max_{x_n} \psi^m(x_n) = \exp \Big(-\frac{1}{2}\Vert b^m_{\tau} \Vert^2_2 \Big)
\end{split}
\end{align}
From here we continue with elimination on discrete variables.
Our final result is a set of functions $g_N()g_{N-1}(m_j)\dots g_1(\cS_1)$ which, when evaluated, gives us the MAP for both continuous $X$ and discrete $M$ variables.

\section{Managing Tractability}\label{sec:tractability}

\subsection{Complexity Analysis}\label{sec:complexity}

In the general case, the \textit{Discrete Bayes Network} formed after elimination has an exponential number of hypotheses~\cite{Murphy02phd}, making it imperative to analyze the complexity of our proposed elimination algorithm.
This analysis allows us to suggest methods which ensure tractability.

Continuous variable elimination has a quadratic cost $O(X^2)$ with $X$ continuous variables, with the continuous variable ordering being vital in maintaining sparsity for efficiency.
When eliminating a variable $a$, the separator $\mathcal{S}_a$ is dependent on the elimination ordering, since a poor ordering can introduce fill-in, leading to a high tree-width (\aka ``dense'' clique connectivity) of the graph.

Discrete variable elimination depends both on the number of discrete variables as well as the number of separators.
For the case of $M$ discrete variables, a tree-width of $W$ for the separator $\mathcal{M}_j$ gives us a cost of $O(K^{W+1})$ on average.
Here $K$ is the cardinality of each discrete variable which we assume to be the same without loss of generality.
In the worst case, the cost of discrete elimination can thus be $O(K^{M})$ \aka exponential.
This exponential cost is thus the primary reason for intractability when performing hybrid inference.

\subsection{Hypothesis Management Techniques}

To ensure tractability of our algorithm, we propose two schemes to manage the exponential hypothesis growth and maintain tractability of our algorithm.
Both these schemes can operate individually or together, leveraging the probabilistic nature of our framework.

%
%

\subsubsection{Hypothesis Pruning}\label{sec:pruning}
To make inference tractable, we adopt hypothesis pruning, a simple yet effective scheme to keep the number of hypotheses manageable.
Given a pruning number $P$, we evaluate the discrete probabilities in our hybrid Bayes network and set the probabilities of the variables outside the top $P$ to $0$. 
Concretely, after we eliminate all the newly added variables, we get a distribution $P(M)$ over the discrete variables $M$, which we can prune to maintain the top $P$ as non-zero.
This reduces the complexity of discrete elimination from $O(K^{W+1})$ to $O(PK)$ (\aka linear complexity) where $K$ is the cardinality of the discrete variables.
Since $P \ll K^W$, we can achieve real-time performance by selecting a suitable value for $P$ without discarding the MAP hypothesis.
Furthermore, the tree representation of the hybrid Bayes network allows for easy implementation.



\subsubsection{Dead Mode Removal}
We can take advantage of the probabilistic nature of our hybrid Bayes network to reduce the number of discrete hypotheses by evaluating the confidence of a discrete mode's assignment.
We compute the marginal probabilities for each discrete mode and if the probability of an assignment is greater than a given threshold value $\DMRThreshold$, we consider that mode ``dead'', due to its low entropy.
Subsequently, we replace that discrete variable in the hybrid Bayes network with the fixed assignment, effectively reducing the discrete probability space.
\begin{equation}\label{eqn:dead-mode-removal}
    P(X_i | M_i \setminus m) =  \underset{m=i}{\text{choose}} P(X_i | M_i); P(m=i) > \DMRThreshold
\end{equation}
where ``choose'' selects the underlying component corresponding to the discrete index $m=i$.

\section{Tackling Nonlinear Problems}

If the function $h^m$ in a factor component is nonlinear, it gives us a \textit{Nonlinear Hybrid Factor} $\HybridNonlinearFactor$, which we need to linearize in order to use in our variable elimination algorithms.
To linearize the factor $\HybridNonlinearFactor$ to a Hybrid Gaussian Factor $\HybridLinFactor$, we apply the Taylor expansion to each component $\HybridGFComponent$ using an initial estimate $X^0_i$.
\begin{align}
    \begin{split}
        & \HybridLinFactor(X_i, M_i) = \Big\{ \HybridComponent(X_i); m \in D(M_i) \Big\} \\
        & \HybridComponent(X_i) \approx \HybridGFComponent(X_i) \\
        &= \frac{1}{\sqrt{|2\pi \Sigma^m|}} \exp \Big( -\frac{1}{2} \Vert H^m \delta x_i - y^m \Vert^2_{\Sigma^m} \Big) \\
        & x_i =  x^0_i + \delta x_i, y^m = z^m - h^m(X^0_i)
    \end{split}
\end{align}
Given that elimination and linearization are independent operations, we can interleave the two, using newer estimates as the latest linearization point and repeating elimination to obtain better estimates.

\section{Experimental Results}\label{sec:experiments}

\subsection{City10000 Dataset}

We showcase our framework on the \textit{City10000} dataset~\cite{Hsiao19icra}, a large scale 2D SLAM problem with ambiguous odometry measurements added at random and uncertain loop closures.
These ambiguities are modeled with hybrid versions of measurement and loop closure factors where the discrete variable is used to resolve the ambiguities. 
These hybrid factors are similar to factor types \#1 and \#3 in~\cite{Hsiao19icra} but our framework is more expressive and not limited to these.

\begin{figure}[h]
    \centering
    \begin{subfigure}{0.23\textwidth}
        \label{fig:city10000_10000_10}
        \includegraphics[width=\textwidth]{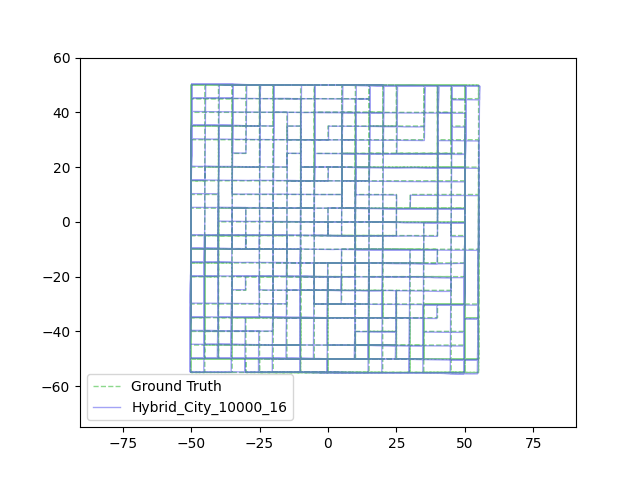}
        \caption{Our hybrid factor graph}
    \end{subfigure}
    \begin{subfigure}{0.23\textwidth}
        \label{fig:city10000_isam2}
        \includegraphics[width=\textwidth]{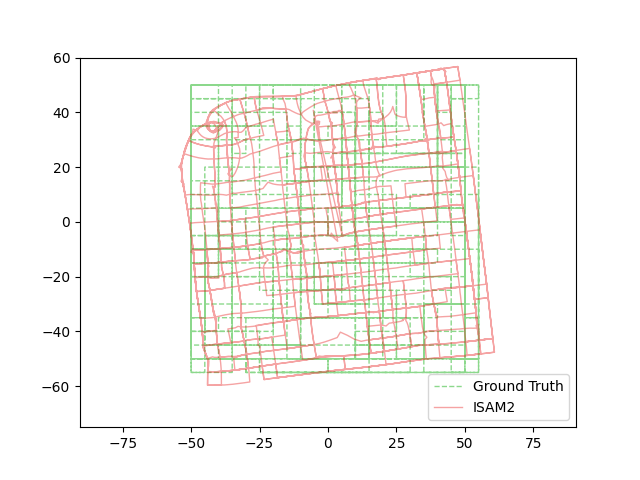}
        \caption{Vanilla iSAM2}
    \end{subfigure}
    \begin{subfigure}{0.23\textwidth}
        \label{fig:city10000_mh_isam2}
        \includegraphics[width=\textwidth]{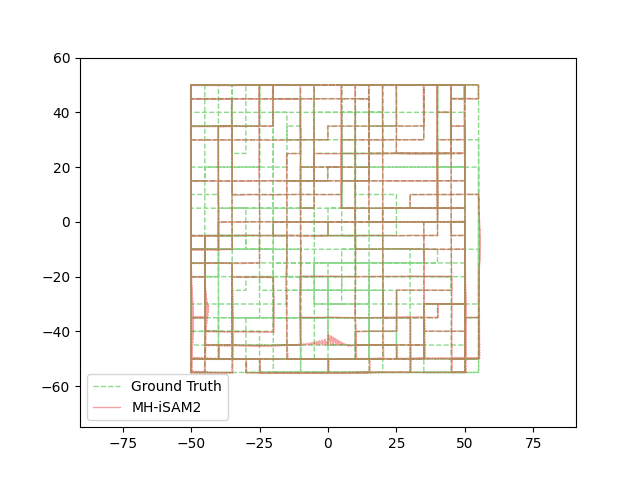}
        \caption{MH-iSAM2}
    \end{subfigure}
        \begin{subfigure}{0.23\textwidth}
        \label{fig:city10000_DCSAM}
        \includegraphics[width=\textwidth]{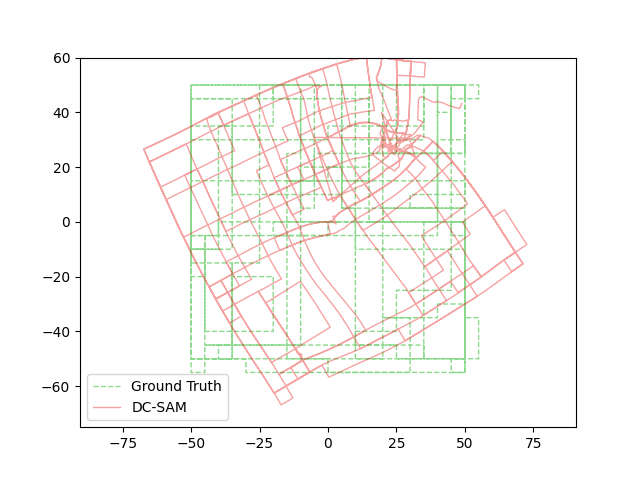}
        \caption{DC-SAM}
    \end{subfigure}
    \caption{Comparative estimates on the City10000 dataset. With 10 hypotheses at each update, our framework handles ambiguous odometry measurements and switching modes for loop closures. iSAM2 sees multiple failure modes. Our framework outperforms MH-iSAM2, while DC-SAM falls into a local minimum at 7000 timesteps and fails to recover.}
    \label{fig:city10000_hybrid}
\end{figure}
To use hybrid factor graphs for estimation on this dataset, we provide a hybrid motion model factor, and a hybrid switching factor for loop closures.
The motion model factor accepts multiple measurements, with the discrete mode selecting which measurement to use.
\begin{equation*}
    \phi(X_k, X_{k+1}, M_k) = 
    \{ \phi_m(X_k, X_{k+1};  \mu_m, \Sigma); m \in |M_k| \}
\end{equation*}
The switching factor has a binary mode where $L_k = 1$ indicates a loop and $L_k = 0$ is not a loop. We achieve this by using a noise model with a tight covariance (same as the motion model) for $L_k = 1$ and a loose covariance of $\Sigma_0 = 10.0$ for non-loops $L_k = 0$.
\begin{equation*}
    \phi(X_k, X_j, L_k) = 
    \{ \phi_l(X_k, X_j;  \mu, \Sigma_l); l \in |L_k| \}
\end{equation*}
We run hybrid elimination after every third hybrid factor, regardless of the number of continuous pose constraints in-between.
After every 10th hybrid elimination, we restrict all non-linear factors according to the fixed values obtained from dead mode removal, and then perform a full batch elimination.
This is fast since there are typically only few discrete variables left after our pruning/DMR strategy.

As baselines, we compare our framework against iSAM2~\cite{Kaess12ijrr}, MH-iSAM2~\cite{Hsiao19icra}, and DC-SAM~\cite{Doherty22ral}. iSAM2 only supports continuous variables so we randomly select one of the multiple ambiguous odometry measurements and provide true loop closures, whereas MH-iSAM2 and DC-SAM are hybrid frameworks capable of both discrete and continuous estimation.
Our approach is able to estimate the full pose graph effectively.
In contrast, MH-iSAM2 displays minor errors which are never resolved, and DC-SAM falls into a local minima which it does not recover from.

\begin{figure}[ht]
    \centering
    \includegraphics[width=0.5\textwidth]{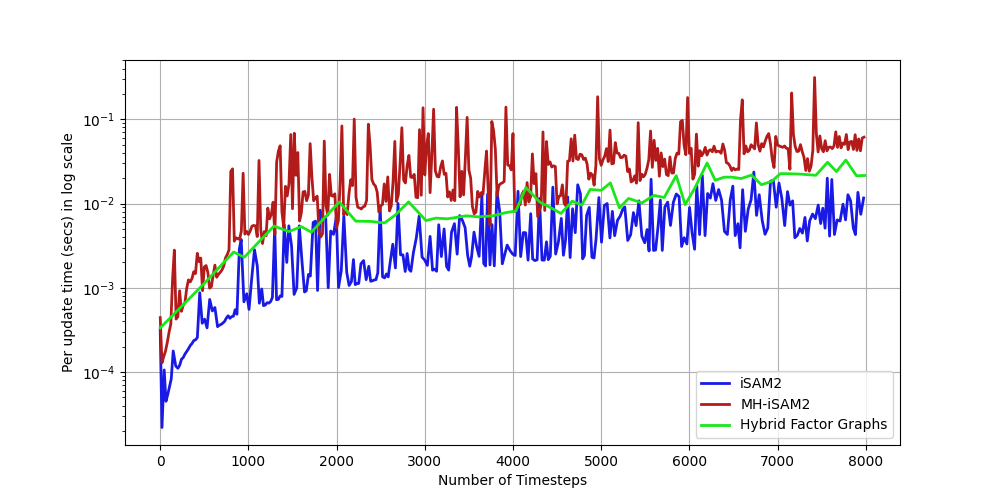}
    \caption{Time per solver update step. Our framework is more efficient than MH-iSAM2~\cite{Hsiao19icra} despite not possesing iSAM2's incremental design features.}
    \label{fig:per_update_time}
\end{figure}

Additionally, we compare runtimes between our framework with pruning $P=10$ and dead mode removal (DMR) $\DMRThreshold = 0.8$, MH-iSAM2, and iSAM2 in Fig.~\ref{fig:per_update_time}.
Our approach is faster than MH-iSAM2 while expectedly slower than the continuous-only iSAM2.

\subsection{Pose Graph Optimization}

We demonstrate our framework on a Pose Graph Optimization (PGO) problem with ambiguous edge measurements.
Following \cite{Doherty22ral}, we are given a pose graph $\mathcal{G} = \{\mathcal{V}, \mathcal{E}\}$ where nodes correspond to poses to be estimated and edges correspond to the noisy measurements between them.
The edges are divided into two sets, $\mathcal{E}_O$ is the set of trusted odometry measurements, $\mathcal{E}_L$ is the set of ambiguous loop closure measurements.
Within $\mathcal{E}_L$, some measurements are drawn from an outlier process.
Thus the problem to solve can be stated as:
\begin{equation}
    \min_{x_i \in \SEThree} \sum_{\{i, j\} \in \mathcal{E}_O} \Vert r_{ij}(x_i, x_j) \Vert^2_\Sigma + \sum_{\{i, j\} \in \mathcal{E}_L} e_{ij}(x_i, x_j, d_{ij})
\end{equation}
where $r_{ij} = \textrm{log}(\bar{x}^{-1}_{ij}x^{-1}_i x_j)^{\vee}$ is the residual function between the measurement $\bar{x}_{ij}$ and the estimated poses in the tangent space, $d_{ij} \in \{0, 1\}$ is the discrete variable indicating the measurement is an inlier or outlier respectively, and
\begin{equation}
    e_{ij}(x_i, x_j, d_{ij}) = \begin{cases}
        -\textrm{log}w_0 + \Vert r_{ij}(x_i, x_j) \Vert^2_{\Sigma}, & d_{ij} = 0 \\
        -\textrm{log}w_1 + \Vert r_{ij}(x_i, x_j) \Vert^2_{\tilde{\Sigma}}, & d_{ij} = 1 \\
    \end{cases}
\end{equation}
with $w_0, w_1 \in [0, 1]$ as prior weights on the inlier and outlier hypotheses and $\tilde{\Sigma}$ the covariance of the outlier process.
\begin{figure}[h]
    \centering
    \begin{subfigure}{0.23\textwidth}
        \label{fig:pgo_CSAIL_rotation}
        \includegraphics[width=\textwidth]{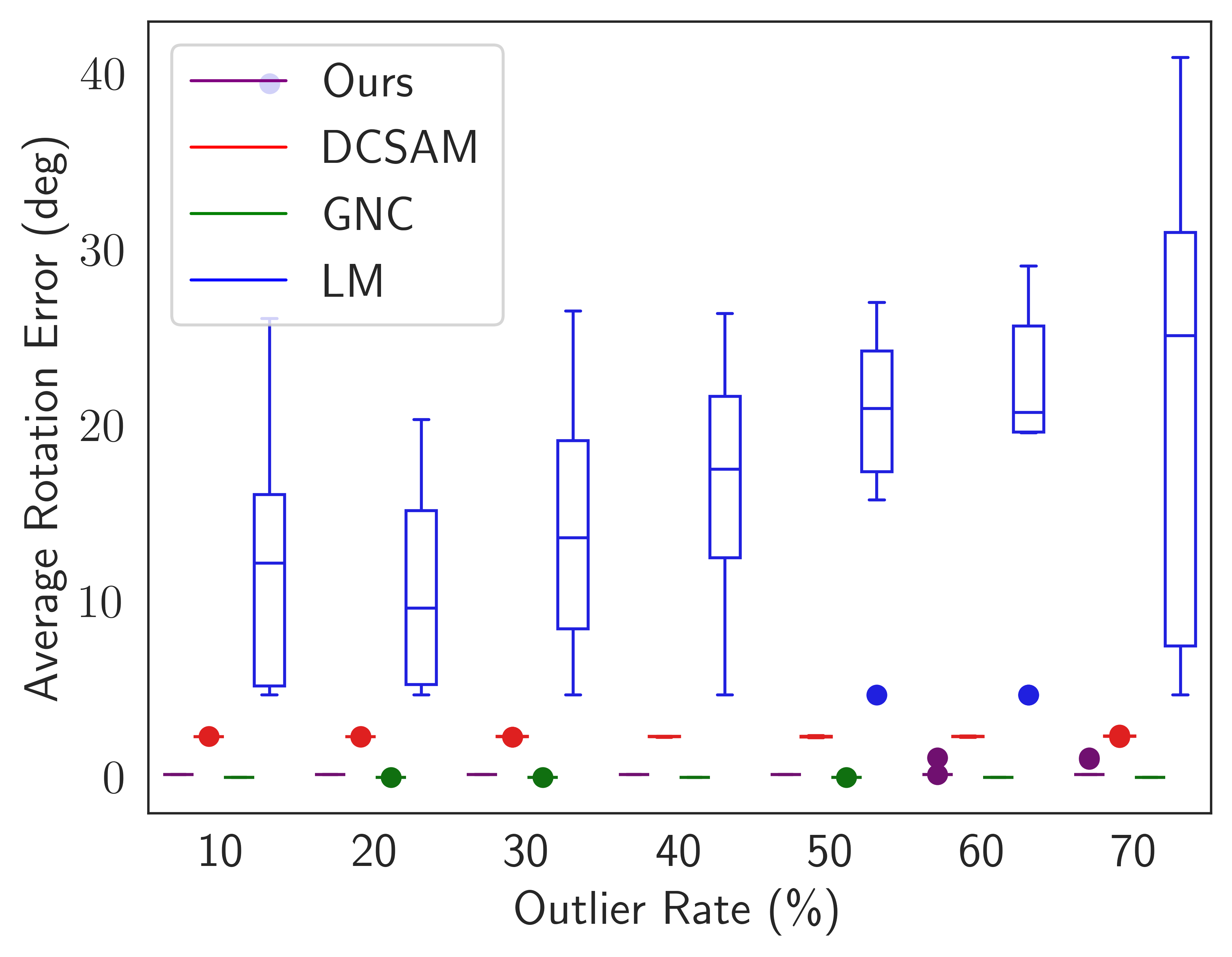}
        \includegraphics[width=\textwidth]{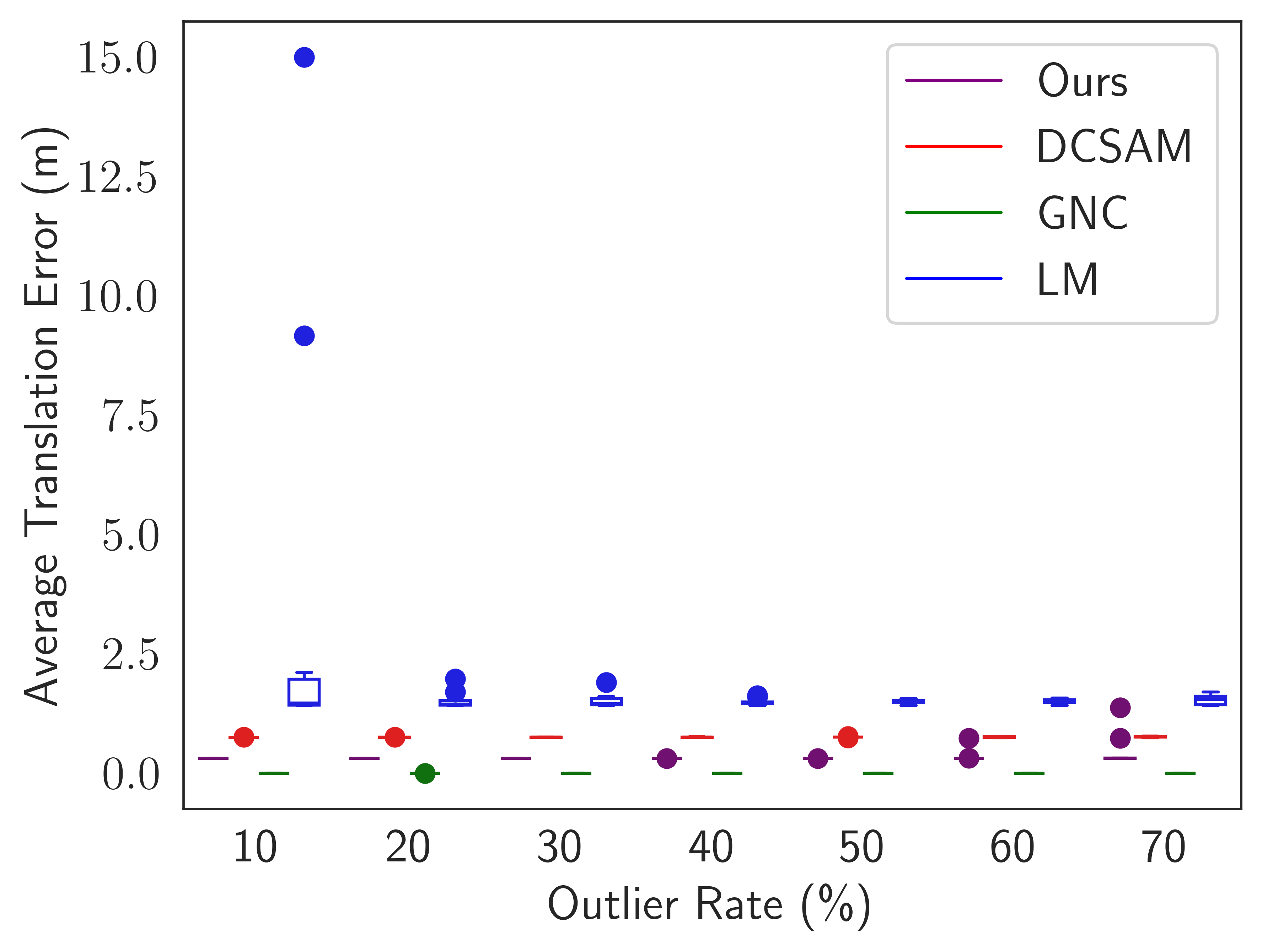}
        \caption{CSAIL}
    \end{subfigure}
    \begin{subfigure}{0.23\textwidth}
        \label{fig:pgo_Intel_rotation}
        \includegraphics[width=\textwidth]{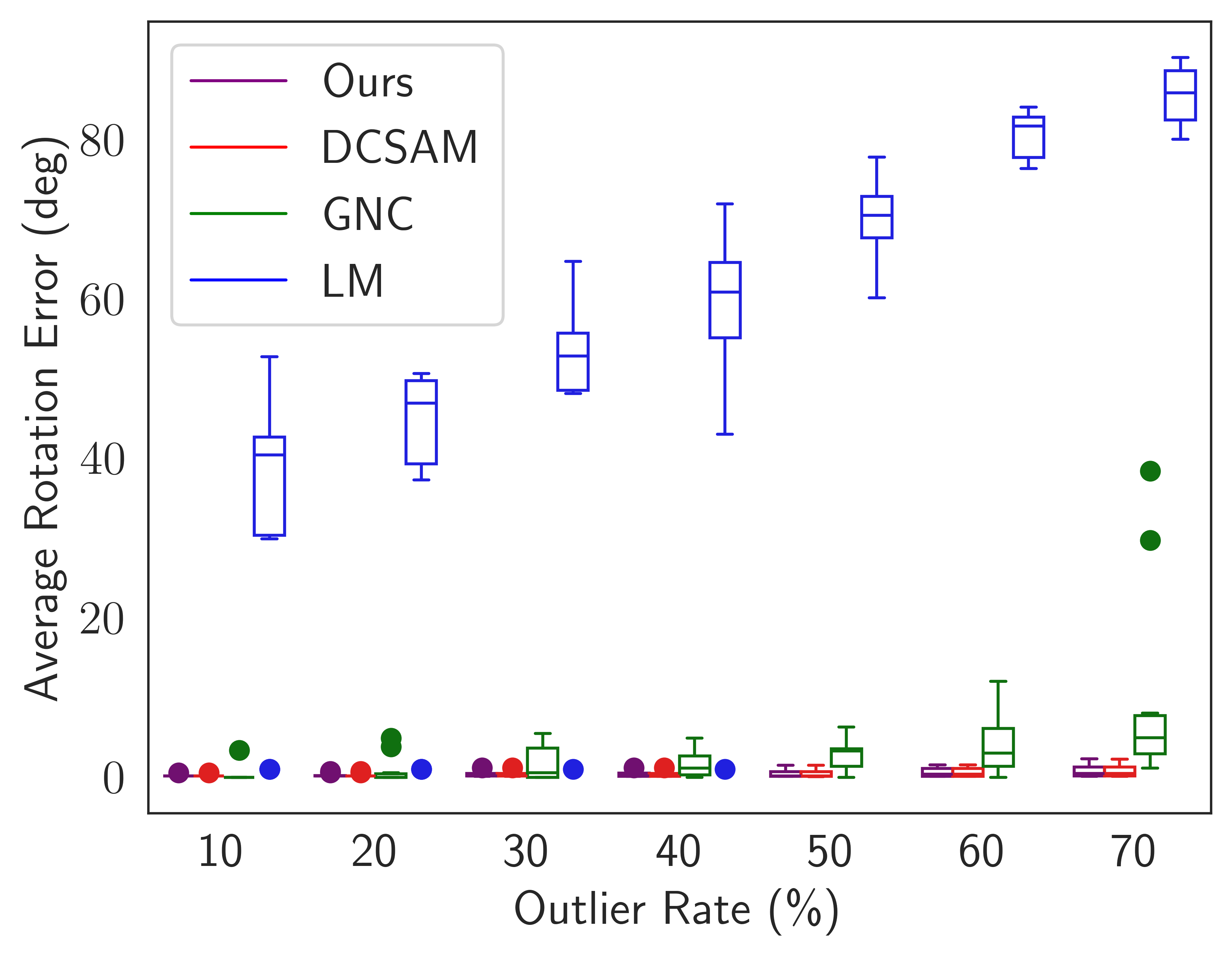}
        \includegraphics[width=\textwidth]{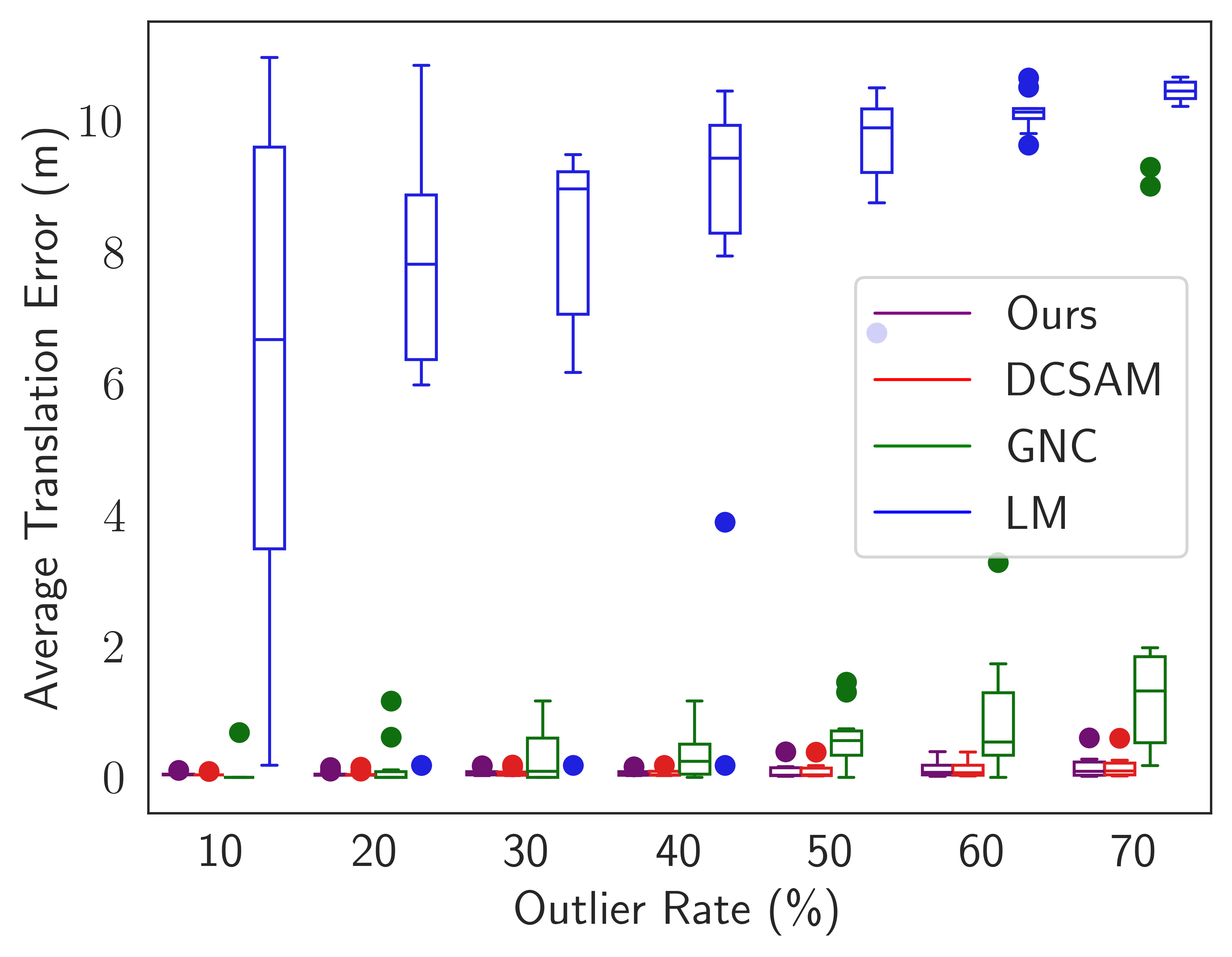}
        \caption{Intel}
    \end{subfigure}
    \caption{Average trajectory errors on the \textit{CSAIL} and \textit{Intel} datasets over 10 Monte Carlo trials. Top row shows the rotation error and bottom row is the translation error.}
    \label{fig:pgo}
    \vspace{-12pt}
\end{figure}

To run the experiments, we use the open source implementation provided by \cite{Doherty22ral}, adding our hybrid factor graph elimination algorithm as one of the methods, along with DCSAM, the GNC optimizer~\cite{Yang20ral_gnc} (which is robust to ouliers), and the Levenberg-Marquadt optimizer as a non-hybrid baseline.
We evaluate our framework on the \textit{CSAIL} and \textit{Intel} datasets which are based on real world floorplans.
The results are summarized in Fig.~\ref{fig:pgo}, and we see that our framework performs at least as well as the other approaches.
\section{Conclusion} \label{sec:conclusion}

In this work, we have presented an algorithm for variable elimination in hybrid factor graphs. The variable elimination algorithm follows a similar paradigm to elimination in continuous/discrete factor graphs, making it applicable to a wide range of robotics problems and straightforward to implement.
We have provided a rigorous derivation of the Sum-Product and Max-Product algorithms for hybrid factor graphs, showing how our formulation allows for exact inference.
Furthermore, we propose Tree Pruning and Dead Mode Removal, two schemes to tackle the exponential growth of discrete hypotheses which allows for efficient performance.

Our experimental results demonstrate the applicability of our approach on a SLAM problem with ambiguous measurements.
On the City10000 dataset, our framework provides a natural way to model the hybrid nature of the problem, and our variable elimination algorithm is able to efficiently compute the MAP estimate better than existing approaches such as MH-iSAM2.

Given the generality of our approach, we hope to further extend it to incorporate various accuracy, robustness, and efficiency improvements proposed over the years towards factor graph based smoothing problems.
Additionally, we hope to develop an incremental solver for hybrid problems in the same vein as iSAM2~\cite{Kaess12ijrr}.
We hope this will lead to widespread applicability and adoption of our work.


\bibliographystyle{IEEEtran}
\bibliography{refs,references}

\end{document}